\ifdefined\pdfoutput\pdfoutput=1\fi
\documentclass[journal]{IEEEtran}

\usepackage[T1]{fontenc}
\usepackage[utf8]{inputenc}
\usepackage{graphicx}
\usepackage{booktabs}
\usepackage{amsmath}
\usepackage{amssymb}
\usepackage{xcolor}

\ifdefined\DeclareUnicodeCharacter
\DeclareUnicodeCharacter{00B0}{\ensuremath{^\circ}}
\fi

\newcommand{\DesignCells}{30}
\newcommand{\GFCells}{18}
\newcommand{\GFXVLAK}{76}
\newcommand{\GFXVLAN}{360}
\newcommand{\GFXVLAPct}{21}

\newcommand{\GFVAAfrozenK}{230}
\newcommand{\GFVAAfrozenN}{360}
\newcommand{\GFVAAfrozenPct}{64}
\newcommand{\GFVAAK}{268}
\newcommand{\GFVAAN}{360}
\newcommand{\GFVAAPct}{74}

\newcommand{\XcellXVLAK}{0}

\newcommand{\XcellGRTNK}{10}

\newcommand{\XcellVAAfrozenK}{44}
\newcommand{\XcellVAAfrozenN}{100}
\newcommand{\XcellVAAK}{68}
\newcommand{\XcellVAAN}{100}

\title{Self-Calibrating Vision-Agent-Action:
Deploying Pick-and-Place to New Robot Cells
without Demonstrations or Fine-Tuning}

\author{Jae Hun Hwang$^{1,2}$, Yechan Seo$^{1,3}$, and Hong-In Won$^{1,4}$%
\thanks{$^{1}$Manufacturing AI Research Center, Korea Institute of Industrial Technology (KITECH), Incheon, Republic of Korea.}%
\thanks{$^{2}$Department of Artificial Intelligence Convergence, Hanyang University, Seoul, Republic of Korea.}%
\thanks{$^{3}$Department of Intelligent Robotics, Sungkyunkwan University, Suwon, Republic of Korea.}%
\thanks{$^{4}$HYU-KITECH Joint Department, Graduate School, Hanyang University, Seoul, Republic of Korea.}%
\thanks{Corresponding author: Hong-In Won (luvhayym@kitech.re.kr).}}

\begin{document}

\maketitle

\begin{abstract}
In simulation, Vision-Agent-Action (VAA) performs pick-and-place
in robot cells with gripper or base-frame changes without robot
demonstrations or fine-tuning. It uses pre-trained vision-language
and language models for interpretation, measures the command map
and camera geometry from its own motion and fingertips, estimates
lens distortion, and plans grasps from measured object depth.
A controlled study of eight public LIBERO-fine-tuned
vision-language-action models (VLAs) varies robot hardware,
camera geometry, appearance, and base frame. Gripper and frame
changes dominate their losses in the tested fractional design.
Across \GFCells{} cells with these changes, frozen VAA achieves
\GFVAAfrozenK/\GFVAAfrozenN{} successes (\GFVAAfrozenPct\%),
the refined VAA \GFVAAK/\GFVAAN{} (\GFVAAPct\%), and the
strongest released VLA on this subset \GFXVLAK/\GFXVLAN{}
(\GFXVLAPct\%). Paired ablations support the self-calibration
bundle, lens estimation under compounded changes, and the
measured-depth pipeline. On a two-step LIBERO-Goal chain with
object state carried between instructions, VAA completes 54/60
chains in changed cells, while four evaluated VLAs complete none.
\end{abstract}


\section{Introduction}

Moving a manipulation system to another robot cell changes more
than its images. A different gripper changes fingertip geometry
and object contact. A rotated command frame changes the direction
in which an action moves the tool center point (TCP).
Camera pose, lens, lighting, support surfaces, and objects may
also change. Collecting demonstrations and fine-tuning a policy
for each cell adds deployment work.

\begin{figure}[t]
    \centering
    \includegraphics[width=\columnwidth]{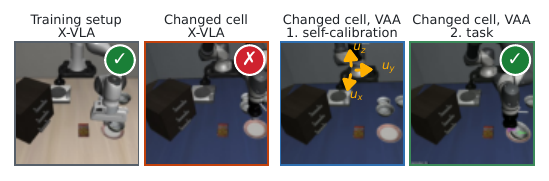}\\[2pt]
    \includegraphics[width=\columnwidth]{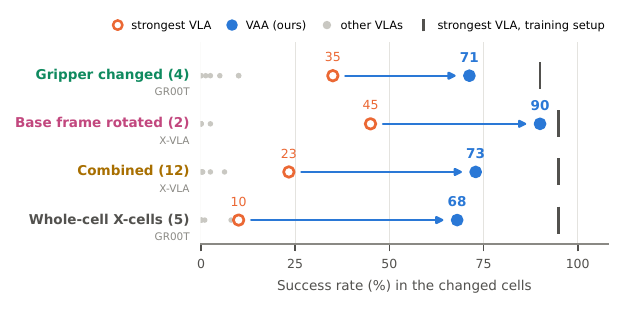}
    \caption{\textbf{VAA retains pick-and-place success under
    tested gripper and base-frame changes that sharply reduce
    released VLA performance.} Top: X-VLA succeeds in the
    training setup and fails in a changed cell; VAA
    self-calibrates in the new cell and completes the task.
    Green and red badges report episode-level success and
    failure. Orange
    arrows show fixed-length image projections of the base
    directions assigned to command channels $u_x$, $u_y$,
    $u_z$ by the measured command map
    (Fig.~\ref{fig:overview}). Bottom: success of the
    strongest VLA in each group (open) and of VAA (filled;
    Tables~\ref{tab:main} and~\ref{tab:xcell}); grey dots show
    the other VLAs. Each tick marks the training-setup success
    of that group's strongest VLA.}
    \label{fig:teaser}
\end{figure}

Public vision-language-action models (VLAs) fine-tuned on LIBERO
can perform tasks in their training setup. Transfer to a new cell
also requires their commands to produce the intended physical
motion under changed hardware and coordinate frames.
Fig.~\ref{fig:teaser} illustrates this deployment problem.

We first examine where released VLAs lose success in a controlled
cell design. Gripper and base-frame changes dominate their
losses in the tested fraction. Both changes alter the relation
between commands and fingertip or object motion. This diagnosis
motivates measuring the new cell's geometry before manipulation.

\begin{figure*}[t]
    \centering
    \includegraphics[width=\textwidth]{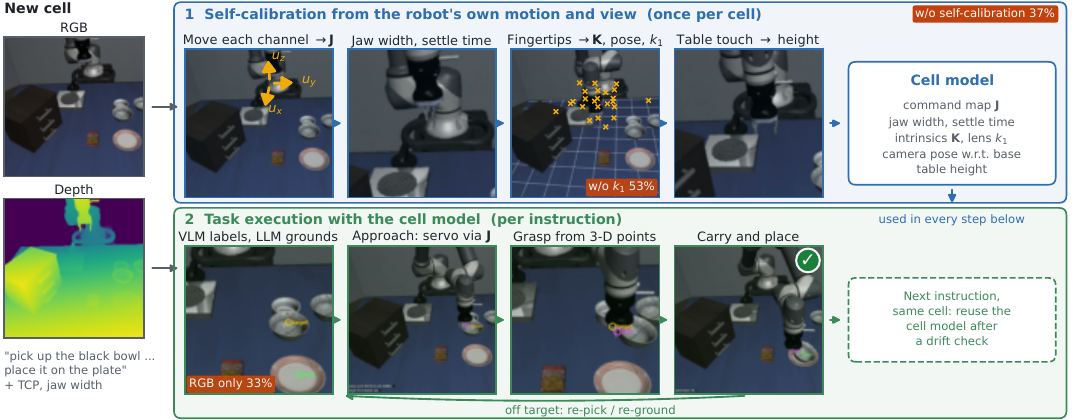}
    \caption{\textbf{A measured cell model connects
    self-calibration to task execution.} Upper lane: in a
    development cell with an RB5 arm, Hand-E gripper, cloth
    table, and moved camera, VAA measures command map
    $\mathbf{J}$, jaw width, settling time, table height,
    and camera geometry. The camera fit uses 24 fingertip
    detections (orange crosses) to estimate intrinsics
    $\mathbf{K}$, pose $[\mathbf{R}\,|\,\mathbf{t}]$, and
    lens term $k_1$. The 10-cm grid is projected at the table
    height obtained in the later touch stage. Command arrows
    use the fixed-length base-direction projections defined
    in Fig.~\ref{fig:teaser}. Lower lane: logged episode
    frames show grounding, servoed approach, grasping from
    object 3-D points, and placement; later instructions
    reuse the model after a drift check. Orange tags give
    ablation success rates out of 60 episodes, versus
    50/60 (83\%) for the full system
    (Section~\ref{sec:ablations}).}
    \label{fig:overview}
\end{figure*}

Vision-Agent-Action (VAA) performs this measurement and executes
pick-and-place without robot demonstrations or fine-tuning.
It combines pre-trained VLM/LLM components with self-calibration.
Before the task, the robot observes its own motion and fingertips
to establish a cell model. During the task, language models
identify the relevant objects, while geometric modules determine
where to grasp and where the TCP must move.
Fig.~\ref{fig:overview} separates these stages.

The study contributes:
\begin{itemize}
    \item A controlled diagnosis identifying gripper and
    command-frame changes as dominant VLA transfer factors
    in the tested design.
    \item A self-calibrating pipeline combining command and
    camera measurement, lens estimation, and measured-depth
    grasp planning.
    \item Paired ablations and two-step manipulation testing
    the roles of cell measurement and calibration reuse.
\end{itemize}

\section{Related Work}

VLAs learn instruction-conditioned actions from demonstrations
and combine visual, linguistic, and motor representations
across tasks and embodiments~\cite{rt2,openvla,pi0,pi05,smolvla,xvla,groot,vlajepa}.
Robustness evaluations examine their sensitivity to visual,
physical, and task changes, including dependence on demonstrated
trajectories~\cite{liberoplus,liberopro,colosseum,intact}.
Camera- and embodiment-transfer methods address parts of this
deployment problem through viewpoint adaptation or cross-embodiment
representations~\cite{anycamvla,camvla,mirage,shadow}.
Our study focuses on gripper and command-frame changes and tests
explicit measurement of the new cell as a deployment procedure.

Foundation-model agents use language, vision, code, and geometric
tools to coordinate manipulation~\cite{cap,voxposer,rekep,okrobot,capx,openrua,waa,rax}.
VAA similarly separates semantic interpretation from geometric
execution, grounding object choices in measured scene geometry.
RoboArchitect~\cite{roboarchitect} uses a coding model to evolve
a harness around a frozen VLA over optimization episodes and
is supplied with the pixel-to-base projection and action
specification. VAA measures the command map, camera geometry,
and lens of each new cell with the same code in every cell; its
per-cell adaptation uses no task-success signal and no code
modification.

Hand--eye calibration, autonomous camera calibration, and visual
self-modeling estimate relationships among robot motion, observed
geometry, and coordinate frames~\cite{tsai,easyhec,kalib,arccalib,selfmodel}.
VAA uses its fingertips as observed features and additionally
measures command-channel motion and gripper behavior.
Its evaluation connects these measurements to pick-and-place
transfer under a parallel-jaw model.

\section{Controlled Study: Where Do VLAs Break?}
\label{sec:study}

\subsection{Tasks, Factors, and Baselines}

All experiments are in simulation. We use
LIBERO-spatial~\cite{libero}, with 10 tasks and
2 initial states per task, giving 20 episodes per cell.
Every method receives the same episodes. The controlled study
contains \DesignCells{} changed design cells.

The factors are arm (A: Panda to UR5e), gripper
(B: Panda to Hand-E), camera pose (C: moved 10\,cm and rotated
$10^\circ$), lens (D: $k_1=0.25$), light
(E: dim warm illumination), table (F: blue cloth),
objects (G: red bowl and blue plate), and base frame
(H: $45^\circ$ rotation). Single-factor cells include additional
levels: RB5 and Doosan arms; PGC, 2FG7, and Schunk grippers;
a high camera; and a $90^\circ$ frame rotation.

Combined changes follow a $2^{8-4}$ resolution-IV fraction:
\begin{equation}
    E=BCD,\quad F=ACD,\quad G=ABC,\quad H=ABD.
    \label{eq:fraction}
\end{equation}
The fraction contains 15 changed cells and a nominal reference.
Changed-cell totals exclude the nominal reference.
In a resolution-IV fraction, main effects are not aliased
with two-factor interactions, but their interpretation assumes
negligible higher-order interactions; two-factor interactions
remain aliased with one another.
Five additional changed development cells provide whole-cell
stress tests.

The baselines are X-VLA, $\pi$0.5, $\pi$0, GR00T N1.7,
OpenVLA, SmolVLA, $\tau$0, and
VLA-JEPA~\cite{xvla,pi05,pi0,groot,openvla,smolvla,tau0,vlajepa}.
We evaluate each released system with its own model, training,
action interface, and authors' inference stack. VAA uses a
third-person RGB-D camera and proprioception; the VLAs receive
RGB images without measured depth. The RGB-only VAA ablation
examines this sensor difference.

\subsection{Evaluation Protocol}

Each baseline uses one seed per evaluation run. Reference runs
of the same setup yield 14/20 and 19/20 successes for OpenVLA,
and 16/20 and 12/20 for SmolVLA.

Frozen VAA was fixed before the design-cell evaluations.
Execution refinements in VAA were selected using the training
setup and development cells, with frozen design outcomes
available. Both versions are reported on the gripper/frame
subset selected from the factor diagnosis.
For the chain evaluation, VAA was not developed on LIBERO-Goal,
and the chain comprises the two most reliable pick-and-place
steps in its shared scene.

\subsection{Factor Diagnosis}
\label{sec:diagnosis}

\begin{figure}[t]
    \centering
    \includegraphics[width=\columnwidth]{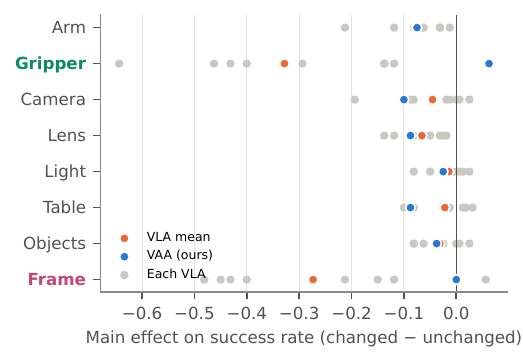}
    \caption{\textbf{Gripper and base-frame changes dominate
    the estimated VLA losses in the tested fraction.}
    Estimated main effects in the resolution-IV fractional
    design; VAA's effects are comparatively small.}
    \label{fig:effects}
\end{figure}

\begin{figure*}[t]
    \centering
    \includegraphics[width=\textwidth]{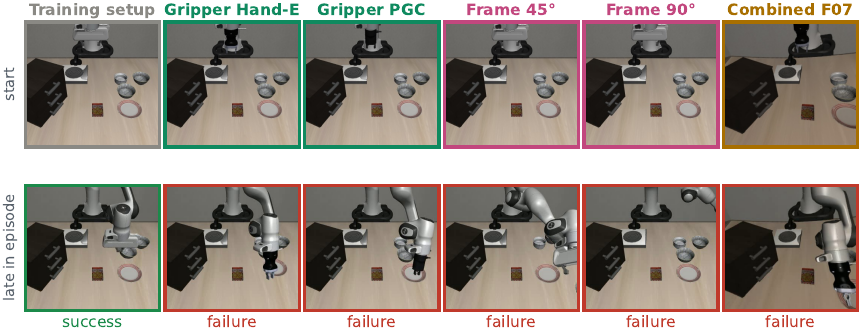}
    \caption{\textbf{$\pi$0.5 succeeds in the training setup
    and fails in five gripper/frame design cells.}
    LIBERO-spatial task~0, initial state~0, one rollout per
    cell. Top: first frame (a base-frame rotation changes the
    command frame and is not visible in the image). Bottom: a
    frame at 70\% of each episode (50\% in the 90$^\circ$
    frame cell, where the arm has left the view by 70\%);
    border colour indicates the outcome.}
    \label{fig:vlacells}
\end{figure*}

Fig.~\ref{fig:effects} identifies gripper and base-frame changes
as the dominant VLA losses in this fraction, and
Fig.~\ref{fig:vlacells} shows one rollout per cell for the
same task and initial state. The remaining evaluation focuses on cells that
change the gripper or rotate the base frame.

In cells that change neither the gripper nor the frame, several
released VLAs retain high aggregate success, with performance
varying by model.

Both dominant factors change the relation between a command and
the resulting motion of the fingers or object. Measuring that
relation through controlled motion provides the cell-specific
information used by VAA.

\section{Vision-Agent-Action}
\label{sec:method}

\subsection{Self-Calibration of the Cell Model}

\begin{figure*}[t]
    \centering
    \includegraphics[width=\textwidth]{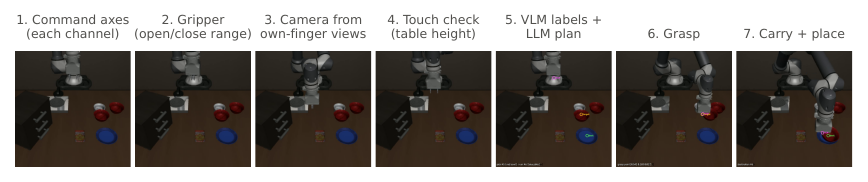}
    \caption{\textbf{The same stages in a second changed
    cell.} VAA in development cell X5 (RB5 arm, OnRobot 2FG7
    gripper, base frame rotated 90$^\circ$, dim light, red
    bowl and blue plate), LIBERO-spatial task~0; one frame
    per stage from the episode log. Steps 1--4 measure the cell; steps 5--7
    perform the instruction with the measured model.}
    \label{fig:pipeline}
\end{figure*}

VAA measures command effects, jaw behavior, and camera geometry
before task execution (Figs.~\ref{fig:overview}
and~\ref{fig:pipeline}). These measurements replace assumptions
inherited from the training setup.

\subsubsection{Command map and jaw behavior}

The robot commands small motions through the available action
channels and measures the resulting TCP displacement:
\begin{equation}
    \Delta\mathbf{x}_{\mathrm{TCP}}
    \approx \mathbf{J}\Delta\mathbf{u},
    \label{eq:command-map}
\end{equation}
where $\Delta\mathbf{u}$ is an action-channel command and
$\mathbf{J}$ is its measured local motion map. In the figures,
command channels 0--2 are assigned to base $+x$/$+y$/$+z$
directions through $\mathbf{J}$.
The controller uses this map to interpret desired TCP
displacements in the current cell.

VAA also measures open and closed fingertip widths and the
settling time after a gripper command. The usable opening
constrains grasp selection, while settling time determines
when to observe the commanded jaw state. The cell model uses
parallel-jaw closure.

\subsubsection{Camera and lens geometry}

The robot observes its fingertips at multiple poses and
associates these observations with its motion and
proprioception to estimate camera intrinsics $\mathbf{K}$
(focal length and principal point) and pose relative to the
robot base $[\mathbf{R}\,|\,\mathbf{t}]$. The resulting
geometry maps base-frame coordinates to image pixels.

VAA estimates radial distortion $k_1$ using table flatness
as a constraint. Geometric errors from a changed lens can
compound with camera and hardware changes even when the object
remains recognizable.

VAA also uses workspace exploration views, a table-height
touch check, and real-time re-reading.
Section~\ref{sec:ablations} evaluates these procedures separately.

\subsection{Grounding and Depth-Based Grasp Planning}

RGB-D observations provide object proposals. A pre-trained
VLM (Qwen3.8, 27B) labels the proposals, and a pre-trained
text LLM (Qwen3, 30B instruct) interprets the instruction to
select a target and destination; both run locally without
fine-tuning.
A geometric resolver and size checks connect those semantic
choices to measured scene entities.

The grasp planner uses the target's own three-dimensional
points to choose where the fingers should approach and close.
The measured jaw opening constrains feasible grasp geometry,
and the cell model connects the selected grasp to a TCP goal.

\subsection{Execution and Verification}

A closed-loop servo executes motion through the calibrated
command map. Verification checks whether placement succeeded.
After an unsuccessful attempt, VAA diagnoses the observed
outcome and can retry.

After initial calibration, a short drift check determines
whether to reuse the cell model or recalibrate, separating
cell preparation from execution of each instruction.

\section{Transfer to Changed Cells}
\label{sec:transfer}

\subsection{Gripper and Base-Frame Changes}

\begin{table*}[t]
    \caption{On the gripper/frame design cells, VAA achieves
    74\% success versus 21\% for the strongest released VLA
    by aggregate success. Success rates (\%) of the eight
    VLAs, frozen VAA, and VAA; each cell contains 20 episodes. $\Delta$: VAA minus
    the strongest VLA in the row, in percentage points.
    Grey: the training setup, before any change.}
    \label{tab:main}
    \centering
    \resizebox{\textwidth}{!}{\definecolor{cgrip}{HTML}{0F8A5F}\definecolor{cframe}{HTML}{C2477E}\definecolor{ccomb}{HTML}{A86F00}\definecolor{cpos}{HTML}{1B8A4A}\definecolor{cneg}{HTML}{C0392B}\definecolor{cref}{HTML}{8A8984}
\begin{tabular}{lrrrrrrrrrrrr}
\toprule
 & & \multicolumn{8}{c}{Released LIBERO-fine-tuned VLAs} & \multicolumn{2}{c}{VAA (ours)} & \\
\cmidrule(lr){3-10}\cmidrule(lr){11-12}
Change & Cells & X-VLA & $\pi$0.5 & $\pi$0 & GR00T N1.7 & OpenVLA & SmolVLA & $\tau$0 & VLA-JEPA & VAA (frozen) & VAA & $\Delta$ \\
\midrule
\textcolor{cref}{\textit{Training setup (reference)}} & \textcolor{cref}{1} & \textcolor{cref}{95} & \textcolor{cref}{90} & \textcolor{cref}{75} & \textcolor{cref}{90} & \textcolor{cref}{70} & \textcolor{cref}{80} & \textcolor{cref}{90} & \textcolor{cref}{95} & \textcolor{cref}{70} & \textcolor{cref}{85} & \\
\midrule
\textcolor{cgrip}{\textbf{Gripper changed}} & 4 & 2 & 0 & 10 & 35 & 5 & 1 & 10 & 0 & 56 & \textbf{71} & \textcolor{cpos}{$+36$} \\
\textcolor{cframe}{\textbf{Base frame rotated}} & 2 & 45 & 0 & 0 & 0 & 0 & 0 & 0 & 2 & 78 & \textbf{90} & \textcolor{cpos}{$+45$} \\
\textcolor{ccomb}{\textbf{Combined}} (incl. gripper or frame) & 12 & 23 & 0 & 2 & 6 & 0 & 0 & 0 & 0 & 64 & \textbf{73} & \textcolor{cpos}{$+50$} \\
\midrule
\textit{All gripper or frame changes} & 18 & 21 & 0 & 4 & 12 & 1 & 0 & 2 & 0 & 64 & \textbf{74} & \textcolor{cpos}{$+53$} \\
\bottomrule
\end{tabular}
}
\end{table*}

Table~\ref{tab:main} reports the \GFCells{} design cells
that change the gripper or rotate the base frame.
Frozen VAA succeeds in \GFVAAfrozenK/\GFVAAfrozenN{}
episodes (\GFVAAfrozenPct\%). VAA reaches
\GFVAAK/\GFVAAN{} (\GFVAAPct\%), compared with
\GFXVLAK/\GFXVLAN{} (\GFXVLAPct\%) for X-VLA,
the strongest VLA by aggregate success on this subset.

Frozen and refined VAA both exceed the strongest VLA in each
reported group (Table~\ref{tab:main}).

\subsection{Whole-Cell Development Stress Tests}

\begin{table*}[t]
    \caption{Across whole-cell development stress tests, VAA
    achieves 68/100 successes and the strongest VLA achieves
    10/100. Success counts out of 20 per cell; all changed
    development cells include a gripper change. The total
    excludes the training setup (grey). $\Delta$: VAA minus
    the strongest VLA in the row, in success counts.}
    \label{tab:xcell}
    \centering
    \resizebox{\textwidth}{!}{\definecolor{cgrip}{HTML}{0F8A5F}\definecolor{cframe}{HTML}{C2477E}\definecolor{ccomb}{HTML}{A86F00}\definecolor{cpos}{HTML}{1B8A4A}\definecolor{cneg}{HTML}{C0392B}\definecolor{cref}{HTML}{8A8984}
\begin{tabular}{lrrrrrrrrrrr}
\toprule
 & \multicolumn{8}{c}{Released LIBERO-fine-tuned VLAs} & \multicolumn{2}{c}{VAA (ours)} & \\
\cmidrule(lr){2-9}\cmidrule(lr){10-11}
Cell & X-VLA & $\pi$0.5 & $\pi$0 & GR00T N1.7 & OpenVLA & SmolVLA & $\tau$0 & VLA-JEPA & VAA (frozen) & VAA & $\Delta$ \\
\midrule
\textcolor{cref}{\textit{Training setup (reference)}} & \textcolor{cref}{18} & \textcolor{cref}{18} & \textcolor{cref}{13} & \textcolor{cref}{19} & \textcolor{cref}{19} & \textcolor{cref}{12} & \textcolor{cref}{20} & \textcolor{cref}{19} & \textcolor{cref}{15} & \textcolor{cref}{17} & \\
\midrule
X1 RB5, Hand-E, cloth, camera & 0 & 0 & 5 & 6 & 0 & 0 & 0 & 0 & 13 & \textbf{18} & \textcolor{cpos}{$+12$} \\
X2 Doosan, PGC, dim, 45° & 0 & 0 & 0 & 0 & 0 & 0 & 0 & 0 & 10 & \textbf{16} & \textcolor{cpos}{$+16$} \\
X4 Schunk, red bowl, lens & 0 & 0 & 4 & 4 & 0 & 1 & \textbf{8} & 0 & 7 & 7 & \textcolor{cneg}{$-1$} \\
X5 RB5, 2FG7, dim, 90° & 0 & 0 & 0 & 0 & 0 & 0 & 0 & 0 & 5 & \textbf{14} & \textcolor{cpos}{$+14$} \\
X6 Doosan, Hand-E, all changed & 0 & 0 & 0 & 0 & 0 & 0 & 0 & 0 & 9 & \textbf{13} & \textcolor{cpos}{$+13$} \\
\midrule
\textit{Changed cells (total)} & 0/100 & 0/100 & 9/100 & 10/100 & 0/100 & 1/100 & 8/100 & 0/100 & 44/100 & 68/100 & \textcolor{cpos}{$+58$} \\
\bottomrule
\end{tabular}
}
\end{table*}

The development cells combine hardware, geometry, and appearance
changes (Table~\ref{tab:xcell}). Across their changed-cell
episodes, VAA achieves \XcellVAAK/\XcellVAAN{} successes,
and frozen VAA achieves
\XcellVAAfrozenK/\XcellVAAfrozenN{}.
The VLAs range from \XcellXVLAK{} to \XcellGRTNK{}
successes out of \XcellVAAN{}.
The following ablations examine which measurements support
the geometric pipeline under these types of changes.

\section{Which Components Matter?}
\label{sec:ablations}

\begin{table}[t]
    \caption{Paired VAA ablations in the PGC gripper,
    $90^\circ$ frame, and combined F07 cells.
    Each cell contains 20 episodes (60 total).
    $\Delta$ is the change in successes from the full system;
    $p$ values are uncorrected exact McNemar tests.
    The text reports the Holm-adjusted interpretation.}
    \label{tab:ablation}
    \centering
    \resizebox{\columnwidth}{!}{\definecolor{cgrip}{HTML}{0F8A5F}\definecolor{cframe}{HTML}{C2477E}\definecolor{ccomb}{HTML}{A86F00}\definecolor{cpos}{HTML}{1B8A4A}\definecolor{cneg}{HTML}{C0392B}\definecolor{cref}{HTML}{8A8984}
\begin{tabular}{lrrrrrr}
\toprule
Variant & \textcolor{cgrip}{Gripper} & \textcolor{cframe}{Frame 90°} & \textcolor{ccomb}{Comb.\ F07} & Total & $\Delta$ & $p$ \\
\midrule
\textbf{Full system} & 15 & 19 & 16 & 50/60 &  &  \\
\midrule
\multicolumn{7}{l}{\textit{Proposed component removed or replaced}} \\
\quad Training-setup calibration reused & 17 & 0 & 5 & 22/60 & \textcolor{cneg}{$-28$} & $<$0.001 \\
\quad No lens estimate ($k_1{=}0$) & 15 & 16 & 1 & 32/60 & \textcolor{cneg}{$-18$} & $<$0.001 \\
\quad RGB only (monocular depth) & 8 & 11 & 1 & 20/60 & \textcolor{cneg}{$-30$} & $<$0.001 \\
\midrule
\multicolumn{7}{l}{\textit{Reference}} \\
\quad Simulator camera/frame (oracle) & 15 & 16 & 13 & 44/60 & \textcolor{cneg}{$-6$} & 0.07 \\
\midrule
\multicolumn{7}{l}{\textit{Implementation details}} \\
\quad No workspace exploration & 16 & 14 & 13 & 43/60 & \textcolor{cneg}{$-7$} & 0.09 \\
\quad No touch check & 14 & 17 & 13 & 44/60 & \textcolor{cneg}{$-6$} & 0.03 \\
\quad No real-time re-reading & 17 & 18 & 15 & 50/60 & \textcolor{cref}{$0$} & 1.00 \\
\bottomrule
\end{tabular}
}
\end{table}

We compare paired episodes in three cells: the PGC gripper
cell, the $90^\circ$ frame cell, and combined cell F07
(gripper, camera, lens, light), one cell from each group of
Table~\ref{tab:main}, with 20 episodes per cell.
The full system succeeds in 50/60 episodes.
Table~\ref{tab:ablation} reports cell-level counts and exact
McNemar tests; the three main component comparisons remain
significant after Holm correction across all seven comparisons.

Reusing the training-setup calibration replaces the command-map
and camera estimates together. The RGB-only variant substitutes
Depth Anything V2 (Large) for measured depth throughout
calibration, perception, and grasping. The full system reuses
calibration after a drift check; the training-setup-calibration,
simulator-geometry, no-lens-estimate, and
no-workspace-exploration variants calibrate before every episode.

\subsection{Self-Calibration Supports Frame Transfer}

Reusing the training-setup calibration reduces success to
22/60 ($p<0.001$). It produces 0/20 on the frame cell and
5/20 on the combined cell, compared with 19/20 and 16/20
for the full system. These losses are consistent with the
motion errors expected when commands use an outdated frame.

On the gripper cell, the reused training-setup calibration
reaches 17/20, compared with 15/20 for the full system.
The calibration benefit is therefore concentrated in the
frame and combined conditions in this ablation.

Replacing estimated camera/frame geometry with simulator
truth yields 44/60 ($p=0.07$).

\subsection{Lens Estimation Matters under Compounded Changes}

Fixing $k_1=0$ reduces success to 32/60 ($p<0.001$).
Nearly all of the loss occurs in the combined cell F07,
where the lens change compounds with gripper and camera
changes: success falls from 16/20 to 1/20.
The frame cell changes from 19/20 to 16/20 and the gripper
cell is unchanged.

\subsection{Measured Depth Supports Gripper Transfer}

Substituting monocular depth reduces success from 50/60 to
20/60 ($p<0.001$), with losses in every cell. The gripper cell
falls from 15/20 to 8/20. One episode ends during calibration
because no table plane is found. Together with the frame and
combined-cell results in Table~\ref{tab:ablation}, these outcomes
support measured depth throughout the VAA pipeline.

\subsection{Other Execution Components}

Removing workspace exploration, the table-touch check, and
real-time re-reading yields 43/60 ($p=0.09$), 44/60
($p=0.03$), and 50/60 ($p=1.00$), respectively.
The touch-check loss comprises six episodes lost and none
gained, but does not survive Holm correction.

\begin{figure*}[t]
    \centering
    \includegraphics[width=0.86\textwidth]{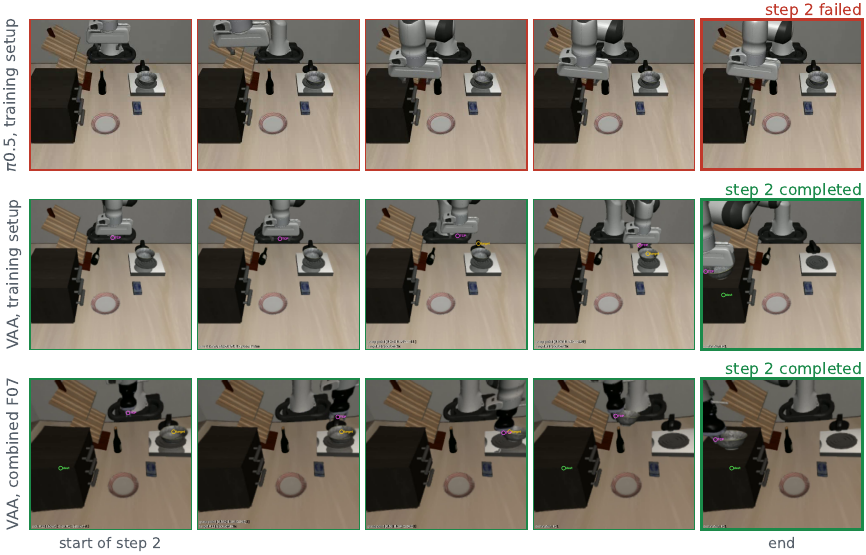}
    \caption{\textbf{VAA grasps the bowl from its carried-over
    position, while $\pi$0.5 moves toward the cabinet without
    grasping it.} Step~2 of the chain moves the bowl from the
    stove to the top of the cabinet; five frames are spread
    over the step. Top: $\pi$0.5, training setup, initial
    state~0. Middle: VAA, same chain. Bottom: VAA in the
    combined cell F07 (gripper, camera, lens, light),
    initial state~1, reusing its calibration. The last
    VAA frames show the bowl still held at the cabinet top,
    before release; scores in Table~\ref{tab:chain}
    use the common release-and-settle test in the text.}
    \label{fig:chainstrip}
\end{figure*}

\section{Calibration Reuse Across Instructions}
\label{sec:reuse}

The two measured initial-calibration costs in the training
setup are 3.3 and 5.7 thousand simulator steps.
Later tasks use a drift check of approximately 90 steps
before reusing the cell model. All preparation costs below
are simulator-step counts rather than wall-clock times.

\begin{table}[t]
    \caption{VAA completes 54 of 60 two-step chains in the
    changed cells; none of the four evaluated VLAs completes
    a chain there. LIBERO-Goal, 20 chains per cell:
    step~1 puts the bowl on the stove, and step~2 moves it
    from there to the top of the cabinet.
    Entries: step~1 successes / completed chains.
    A step counts only after release and the settling
    procedure described in the text.}
    \label{tab:chain}
    \centering
    \resizebox{\columnwidth}{!}{\definecolor{cgrip}{HTML}{0F8A5F}\definecolor{cframe}{HTML}{C2477E}\definecolor{ccomb}{HTML}{A86F00}\definecolor{cpos}{HTML}{1B8A4A}\definecolor{cneg}{HTML}{C0392B}\definecolor{cref}{HTML}{8A8984}
\begin{tabular}{lccccc}
\toprule
 & \multicolumn{4}{c}{Released VLAs} & \\
\cmidrule(lr){2-5}
Cell & SmolVLA & X-VLA & $\pi$0.5 & VLA-JEPA & VAA \\
\midrule
\textcolor{cref}{\textit{Training setup}} & 19 / 0 & 20 / 0 & 20 / 0 & 20 / 0 & 20 / \textbf{16} \\
\textcolor{cgrip}{Gripper PGC} & 0 / 0 & 0 / 0 & 0 / 0 & 0 / 0 & 20 / \textbf{19} \\
\textcolor{cframe}{Frame 90°} & 0 / 0 & 1 / 0 & 0 / 0 & 0 / 0 & 20 / \textbf{19} \\
\textcolor{ccomb}{Combined F07} & 0 / 0 & 0 / 0 & 0 / 0 & 0 / 0 & 17 / \textbf{16} \\
\midrule
\textit{Changed cells} & 0 / 0 & 1 / 0 & 0 / 0 & 0 / 0 & 57 / \textbf{54} \\
\bottomrule
\end{tabular}
}
\end{table}

We evaluate a two-step LIBERO-Goal chain: ``put the bowl on
the stove,'' followed by ``put the bowl on top of the
cabinet.'' Object state carries over between instructions.
The evaluation uses 20 chains per cell, with initializations
0--19, on the training setup and the PGC gripper,
$90^\circ$ frame, and combined F07 cells of
Table~\ref{tab:ablation}.

The same completion rule applies to VAA and the VLAs.
After release, the robot returns to its start pose with
open fingers, the simulator advances 250 physics steps,
and the goal is re-checked. A step counts only if its goal
holds after this settling procedure.

SmolVLA, X-VLA, $\pi$0.5, and VLA-JEPA each achieve
19--20/20 step~1 successes in the training setup, but each
records 0/20 completed step~2 goals. With the bowl on the
stove, they move toward the cabinet without grasping it.
This behavior is consistent with the trajectory-memorization
finding of LIBERO-PRO~\cite{liberopro}
(Fig.~\ref{fig:chainstrip}, top).

Across the changed cells, the four VLAs achieve only one
step~1 success in total, from X-VLA on the $90^\circ$
frame cell, and complete 0/240 chains
(4 VLAs $\times$ 60 chains).
VAA achieves 57/60 step~1 successes and completes
54/60 chains in the changed cells. In the training setup,
it achieves 20/20 step~1 successes and completes
16/20 chains (Table~\ref{tab:chain}).
Paired exact McNemar comparisons of completed chains favor
VAA over each VLA in the changed cells
(54 versus 0, $p=1.1\times10^{-16}$) and in the training
setup (16 versus 0, $p=3.1\times10^{-5}$).

VAA initially calibrates once per cell, using
2.2--6.5 thousand simulator steps, and performs a drift check
of 64--86 steps before each later chain. The check reuses
the cell model in 74 of the 76 later chains and triggers
one recalibration each in the training setup and the
$90^\circ$ frame cell.
Across the training setup and changed cells, VAA has
3 failures at step~1, all in the combined cell F07, and
7 failures at step~2.

\section{Failure Analysis}

\begin{figure}[t]
    \centering
    \includegraphics[width=\columnwidth]{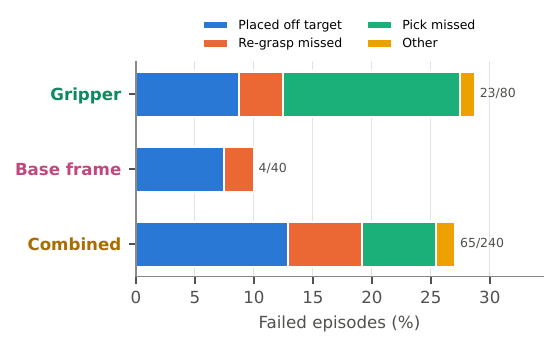}
    \caption{\textbf{Off-target placements and missed picks
    account for most remaining VAA failures.}
    Failure categories for VAA on the \GFCells{}
    gripper/frame cells of Table~\ref{tab:main}, by group.
    Each bar shows the share of all episodes in the group
    that failed at each stage; numbers give failed/total
    episodes.}
    \label{fig:failures}
\end{figure}

On the \GFCells{} gripper/frame cells, VAA has
92 failures: 41 placements remain off target after
correction, 19 re-grasps miss, 27 picks miss, and 5
failures fall into other categories
(Fig.~\ref{fig:failures}). Two tasks, which pick the bowl
from the top drawer of the cabinet and from the cookie box
and place it on the plate,
account for 40\% of failures. Most remaining failures
therefore occur at contact and placement.

An additional single-factor test uses a Robotiq 2F-85 gripper,
whose fingertips close along an arc outside VAA's parallel-jaw
model. VAA achieves 0/20 successes in this cell, compared with
0--6/20 for
the VLAs. This additional cell is excluded from the
\GFCells{}-cell gripper/frame totals.

\section{Conclusion}

VAA connects pre-trained language and vision-language models
to measured cell geometry for pick-and-place.
It calibrates command effects, camera geometry, and lens
distortion, then selects grasps from measured object depth.

On the gripper/frame design cells, frozen VAA achieves
\GFVAAfrozenK/\GFVAAfrozenN{} successes, compared with
\GFXVLAK/\GFXVLAN{} for the strongest VLA on that
subset. The refined VAA reaches \GFVAAK/\GFVAAN{}.
Paired ablations support the calibration bundle for frame
transfer, lens estimation under compounded changes, and the
measured-depth pipeline. With drift-checked calibration reuse
across chained instructions, VAA completes 54/60 two-step
chains in changed cells, while the four evaluated VLAs
complete none.
These simulation results support explicit measurement of the
new cell as a deployment step for the tested gripper and
frame changes.

\section*{Funding}

This work was supported by the Korea Institute of Industrial
Technology (KITECH) under project PEH26020.

\section*{Acknowledgment}

Under the authors' direction, Claude (Anthropic) assisted in
implementing the analysis, simulation and figure-rendering code
and in editing the manuscript text. The authors defined the
research questions, design, analysis and conclusions, and take
full responsibility for the content.

\end{document}